\documentclass[conference]{IEEEtran}
\IEEEoverridecommandlockouts
\usepackage{cite}
\usepackage{amsmath,amssymb}
\usepackage{graphicx}
\usepackage{booktabs}
\usepackage{multirow}
\usepackage{array}
\usepackage{tabularx}
\usepackage{url}
\usepackage{xcolor}
\usepackage[font=small,skip=3pt]{caption}
\usepackage{balance}
\begin{document}
\raggedbottom

\title{SDoH-Aware Narrative Anchoring Bias in Medical LLMs for Trustworthy Clinical Decision Support}

\author
{\IEEEauthorblockN{Ahnaf Atef Choudhury}
\IEEEauthorblockA{\textit{Department of Information Sciences and Technology} \\
\textit{George Mason University, USA}\\
achoudh9@gmu.edu}
\and
\IEEEauthorblockN{Ramkrishna Saha}
\IEEEauthorblockA{\textit{Department of Computer Science} \\
\textit{The University of Texas at Dallas, USA}\\
ramkrishna.saha@utdallas.edu}
}

\maketitle


\begin{abstract}
Medical large language models are often judged by how many clinical questions they answer correctly. That view is useful, but it misses a practical risk. A model may know the right answer and still change its response when the same case is written in a different patient voice. This paper evaluates that risk as SDoH aware narrative anchoring bias. We use NarrativeShield SDoH MedQA, a counterfactual medical question answering dataset in which each case appears in persona based narratives while the answer key remains fixed. The dataset is reshaped from wide format into case grouped persona rows. We evaluate three open source instruction tuned LLMs from the Qwen2.5 family: 1.5B, 3B, and 7B. The final experiment uses 300 clinical cases and produces 8,100 model responses across three prompting conditions. We report persona level accuracy, counterfactual consistency, correct consistency, and narrative sensitivity error. Qwen2.5 7B achieves the best accuracy at 56.33 percent and the best correct consistency at 40.33 percent. Paired McNemar exact tests show significant accuracy gains for 7B over 3B in all prompt settings. Even so, narrative sensitivity remains, with the lowest error still at 31.67 percent. These results suggest that trustworthy clinical decision support should be evaluated by both average correctness and stability across medically equivalent patient narratives.
\end{abstract}

\begin{IEEEkeywords}
medical large language models, social determinants of health, narrative anchoring bias, clinical decision support, trustworthy AI, counterfactual evaluation
\end{IEEEkeywords}

\section{Introduction}
Medical large language models have made rapid progress on clinical question answering. MultiMedQA and Med PaLM showed that instruction tuned models can answer professional medical questions at a level that earlier general language models did not reach \cite{singhal2023clinical}. Still, a clinical support system has to do more than pass exam style questions. The system should act the same way when the clinical facts are identical. It also needs to consider patient context, but not let factors like voice, stress, or social situation distract from the evidence. Recent studies on clinical LLM limitations highlight this issue. Static benchmarks cannot measure every skill needed for safe decision support \cite{kanjee2024limitations}.

One question that has not been explored as much is whether a model gives the same answer when the same case is described in different ways by patients. In reality, patients rarely use a standard format. Some might use medical terms to describe their symptoms, while others may sound unsure. Another patient might talk about money problems or delays in getting care. These details can show things like social factors, health knowledge, barriers to access, or a lack of trust in the healthcare system. Some context can be clinically meaningful. The concern in this paper is different. We ask whether the model changes its answer when the clinical facts and gold label are unchanged.

We call this failure mode SDoH aware narrative anchoring bias. In this setting, anchoring means the model seems to focus on how a patient speaks or the social context, instead of the medical evidence that should guide its answer. This is important because average accuracy can cover up problems. A model might get many persona-level questions right, but still give different answers to similar cases.

This study evaluates that issue using NarrativeShield SDoH MedQA \cite{chugh2026dataset}. The dataset provides persona based variants for medical questions. Unlike prior work that often studies demographic attributes or broad sociodemographic labels, this study focuses on narrative form itself. The central question is whether a model changes its answer when the same clinical case is expressed through different patient voices while the answer key remains unchanged. We evaluate three open source instruction tuned models with matched sampling and deterministic decoding. The goal is not to introduce a new clinical model. Rather, the goal is to test whether answer accuracy and narrative robustness tell the same story.

The study makes four main contributions. First, it operationalizes SDoH aware narrative anchoring bias as a case grouped counterfactual evaluation problem. It then compares model scale using the same 300 clinical cases across three Qwen2.5 model sizes. It reports case grouped consistency metrics that separate stable correct behavior from unstable responses. Finally, it adds bootstrap confidence intervals and paired McNemar exact tests so that the analysis does not rely only on point estimates.

\section{Related Work}
\subsection{Medical Question Answering and Clinical LLMs}
Medical question answering benchmarks are widely used to measure clinical knowledge in language models. MedQA introduced medical exam questions as a challenging open domain QA task \cite{jin2021medqa}. MedMCQA expanded this line of work with more than 194,000 multiple choice questions from medical entrance examinations \cite{pal2022medmcqa}. PubMedQA focused on biomedical research questions built from PubMed abstracts \cite{jin2019pubmedqa}.

Large language models have improved results on these benchmarks. MultiMedQA brought together several medical QA tasks and included consumer health questions using HealthSearchQA \cite{singhal2023clinical}. The study found clear benefits from scaling and instruction tuning. It also highlighted the importance of alignment, safety review, and domain-specific evaluation. More recent clinical work has made a similar argument. High benchmark scores do not guarantee careful reasoning when a model faces changes in context, uncertainty, or presentation \cite{kanjee2024limitations}. A systematic review in \textit{JAMA} also noted that current healthcare LLM evaluations may not fully identify where these systems are most useful or safest \cite{bedi2025jama}. Recent safety effectiveness benchmarking reaches a related conclusion. The CSEDB framework uses clinician consensus metrics and open ended clinical scenarios to examine practical utility beyond medical exam performance \cite{wang2025csedb}.

\subsection{Clinical Language Modeling}
Biomedical and clinical NLP research has long shown that domain adaptation can improve medical text understanding. BioBERT improved biomedical named entity recognition, relation extraction, and biomedical question answering after pretraining on biomedical corpora \cite{lee2020biobert}. GatorTron later scaled clinical language modeling with large biomedical and clinical corpora and improved several clinical NLP tasks \cite{yang2022gatortron}. These models show the value of medical text exposure. They do not, by themselves, answer whether a clinical answer remains stable across socially varied patient narratives.

\subsection{SDoH and Bias in Healthcare AI}
Social determinants of health are often described in free text. Recent work has shown that LLMs can extract SDoH factors from clinical notes across institutions, although performance can vary by setting \cite{keloth2025sdoh}. Guevara and colleagues also showed that LLM based extraction can recover adverse social factors from electronic health records \cite{guevara2024sdoh}. This literature treats SDoH as information that may be useful for care. Our study instead examines a related risk. SDoH grounded narrative cues may also become a source of instability if the model overreacts to patient voice.

Bias studies in medical AI support this concern. Omiye and colleagues showed that LLMs can reproduce race based medical misconceptions \cite{omiye2023race}. Omar and colleagues found that sociodemographic changes can alter LLM medical decisions even when the clinical case remains otherwise matched \cite{zack2025sociodemographic}. Broader fairness work also warns that biased behavior can enter clinical AI systems at many points in the workflow \cite{chen2023fairness,obermeyer2019racial,rajkomar2018fairness}. Prior work therefore gives strong evidence that medical LLMs can be sensitive to demographic and social cues. In our study, we focus on a specific question. It asks whether narrative form alone can change answers when the clinical facts and gold label are fixed. This matters because the way a narrative is presented can vary, even if no demographic details are changed. For example, a patient might come across as more formal, uncertain, distressed, or socially limited, but the clinical answer does not change. We see this as a problem of stability across cases, not just about accuracy at the persona level.

\section{Dataset and Study Design}
This study uses NarrativeShield SDoH MedQA from Hugging Face \cite{chugh2026dataset}. The dataset contains medical multiple choice cases with answer options, a correct answer index, metadata, and persona based narrative variants. The relevant persona columns are \texttt{persona\_alpha}, \texttt{persona\_beta}, and \texttt{persona\_gamma}. Each variant keeps the same answer key while changing the style in which the case is presented.

The dataset is provided in wide format. One row contains the original case and its persona columns. For evaluation, we reshape it into long format. Each \texttt{question\_id} is expanded into one row per persona. The resulting row stores the case identifier, persona label, narrative text, options, gold answer, and metadata. This layout makes the counterfactual comparison direct because the three rows from the same case can be grouped after inference.

The persona fields are used conservatively in this study. Alpha is used as the main case narrative, while beta and gamma represent alternative patient voices from the dataset. We do not give these labels any specific demographic identities. The evaluation assumes that the clinical answer stays the same for all matched variants. Because the dataset provides a fixed answer key across matched persona variants, we treated the variants as counterfactual forms of the same case. We did not perform an additional physician adjudication of clinical equivalence, which is addressed as a limitation.

\begin{figure*}[!t]
\centering
\vspace{-2mm}
\includegraphics[width=\textwidth,trim=0 0 0 0,clip]{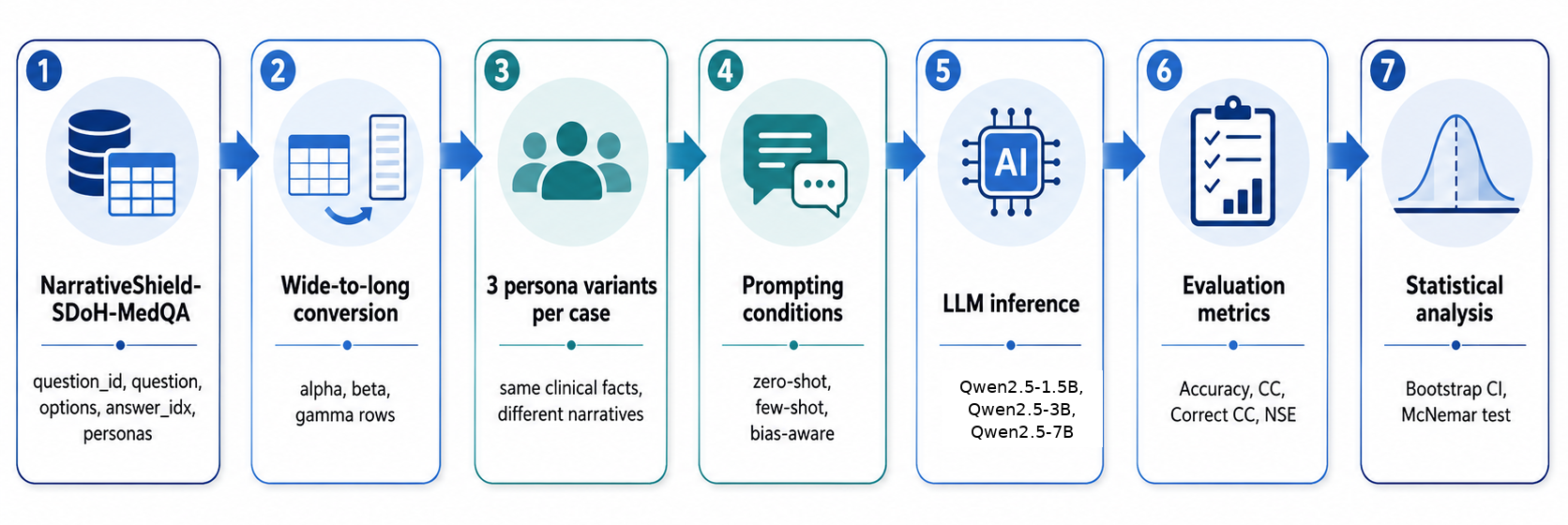}
\vspace{-2mm}
\caption{Counterfactual evaluation pipeline for SDoH aware narrative anchoring bias. The same clinical case is evaluated across persona variants, prompt settings, and model scales. The final analysis separates answer accuracy from case level narrative robustness.}
\vspace{-2mm}
\label{fig:pipeline}
\end{figure*}

For the final experiment, we use 300 unique cases sampled with seed 42. For every sampled case, all persona variants are retained. The sample therefore contains 900 persona level rows. Each model is evaluated under the same prompt settings, which gives 2,700 responses per model. Across all three models, the final analysis uses 8,100 responses. Fig.~\ref{fig:pipeline} summarizes the full pipeline.

\section{Methodology}
\subsection{Prompting Conditions}
We use zero shot, few shot, and bias aware prompting. Table~\ref{tab:prompts} gives the core instruction used in each setting. The full prompt also includes the clinical question, answer choices, and a request for a single option letter. The few shot setting adds two solved examples before the target case.

\begin{table}[!b]
\centering
\footnotesize
\caption{Core Prompt Instructions}
\label{tab:prompts}
\resizebox{\columnwidth}{!}{%
\begin{tabular}{lp{0.66\columnwidth}}
\toprule
Prompt & Core instruction \\
\midrule
Zero shot & Choose the best option and return one letter with a brief explanation. \\
Few shot & Use two solved medical examples before answering the target case. \\
Bias aware & Focus on clinical evidence. Ignore writing style or patient background unless clinically relevant. \\
\bottomrule
\end{tabular}%
}
\end{table}

This prompt comparison is intentionally simple. The purpose is to test whether a lightweight instruction can reduce narrative anchoring. It is not meant to be a full fairness intervention. We keep the prompts brief to focus on how the model responds to different patient narratives, rather than on complex prompt design.

\subsection{Model Selection}
This study looks at Qwen2.5 1.5B Instruct, Qwen2.5 3B Instruct, and Qwen2.5 7B Instruct \cite{qwen2025model}. These are open source, instruction-tuned language models with smaller to mid-sized parameter counts compared to leading clinical systems. We chose them so the experiment could be repeated using typical academic computing resources and without needing paid clinical APIs. Using models from the same family allows us to compare their performance at different scales while keeping the dataset, decoding setup, and prompt design the same. The purpose of this selection was not to provide a broad leaderboard across medical LLMs, but to test whether narrative variation affects answers under a controlled within family scaling setup. Instead, it checks if narrative robustness can be measured in a reproducible way and whether increasing the model size from 1.5B to 7B improves both accuracy and counterfactual robustness.

\subsection{Evaluation Metrics}
Let $M$ denote the number of unique clinical cases and $K$ denote the number of persona variants per case. In this study, $K=3$. Let $y_i$ be the gold answer for case $i$ and $\hat{y}_{i,j}$ be the model prediction for persona variant $j$ of case $i$.

\begin{equation}
Acc=\frac{1}{MK}\sum_{i=1}^{M}\sum_{j=1}^{K}\mathbb{I}(\hat{y}_{i,j}=y_i).
\end{equation}

Counterfactual consistency measures whether all persona variants of a case receive the same predicted answer.

\begin{equation}
CC=\frac{1}{M}\sum_{i=1}^{M}\mathbb{I}
\left(\hat{y}_{i,1}=\hat{y}_{i,2}=\cdots=\hat{y}_{i,K}\right).
\end{equation}

Correct consistency is stricter. It requires every persona variant of a case to be answered correctly.

\begin{equation}
CorrectCC=\frac{1}{M}\sum_{i=1}^{M}\mathbb{I}
\left(\forall j \in \{1,\ldots,K\},\hat{y}_{i,j}=y_i\right).
\end{equation}

Narrative sensitivity error measures cases where at least one persona is correct while another equivalent persona is wrong.

\begin{equation}
NSE=\frac{1}{M}\sum_{i=1}^{M}\mathbb{I}
\left(
\exists a,b \in \{1,\ldots,K\}:
\hat{y}_{i,a}=y_i \land \hat{y}_{i,b}\ne y_i
\right).
\end{equation}

\section{Experimental Setup}
Table~\ref{tab:config} summarizes the experimental configuration. Because all models use the same sampled cases, comparisons are paired at the level of case identifier, persona, and prompt setting.

\begin{table}[t]
\centering
\caption{Experimental Configuration}
\label{tab:config}
\resizebox{\columnwidth}{!}{%
\begin{tabular}{ll}
\toprule
Setting & Value \\
\midrule
Sample size & 300 clinical cases \\
Persona variants & 3 per case \\
Total responses & 8,100 \\
Models & Qwen2.5 1.5B, 3B, and 7B Instruct \\
Prompt settings & Zero shot, few shot, bias aware \\
Decoding & Deterministic with \texttt{do\_sample=False} \\
Max new tokens & 120 \\
Quantization & 4 bit loading with bitsandbytes \\
Hardware & Google Colab Pro with NVIDIA L4 GPU \\
Invalid outputs & Counted as incorrect \\
Statistical tests & Bootstrap CI and McNemar exact test \\
\bottomrule
\end{tabular}%
}
\end{table}

We used deterministic decoding to reduce sampling noise. Each response was parsed into one answer letter. If no valid option could be extracted, the output was treated as invalid and counted as incorrect. Invalid outputs were rare. Qwen2.5 1.5B produced 10 invalid responses out of 2,700 outputs. Qwen2.5 3B produced 22 invalid responses, and Qwen2.5 7B produced 6 invalid responses. All invalid outputs were counted as incorrect.

We report bootstrap 95 percent confidence intervals with 2,000 resamples. McNemar exact tests are used for paired accuracy comparisons. This test is appropriate here because each model sees the same case, persona, and prompt combination.

\section{Results}
\subsection{Main Performance Results}
Table~\ref{tab:main_results} reports the final 300 case results. Accuracy improves with scale across all prompt settings. Under zero shot prompting, accuracy rises from 38.44 percent for Qwen2.5 1.5B to 48.78 percent for Qwen2.5 3B and 56.33 percent for Qwen2.5 7B. The same pattern appears under few shot and bias aware prompting. The 7B model also gives the highest correct consistency, reaching 40.33 percent under zero shot and bias aware prompting.

\begin{table*}[!t]
\centering
\footnotesize
\caption{Main Results on 300 Clinical Cases}
\label{tab:main_results}
\resizebox{0.98\textwidth}{!}{%
\begin{tabular}{llrrrrrr}
\toprule
Model & Prompt & Accuracy & Accuracy 95\% CI & CC & CC 95\% CI & Correct CC & NSE \\
\midrule
Qwen2.5 1.5B & Bias aware & 39.89 & 36.78 to 43.22 & 39.00 & 33.67 to 44.67 & 17.33 & 43.00 \\
Qwen2.5 1.5B & Few shot & 38.67 & 35.67 to 41.89 & 38.67 & 33.00 to 44.33 & 18.00 & 42.33 \\
Qwen2.5 1.5B & Zero shot & 38.44 & 35.33 to 41.67 & 47.00 & 41.00 to 52.67 & 19.67 & 36.67 \\
Qwen2.5 3B & Bias aware & 48.56 & 45.44 to 51.89 & 46.67 & 41.00 to 52.33 & 28.00 & 39.67 \\
Qwen2.5 3B & Few shot & 42.67 & 39.44 to 45.89 & 39.00 & 33.67 to 44.33 & 22.67 & 40.33 \\
Qwen2.5 3B & Zero shot & 48.78 & 45.44 to 52.11 & 45.33 & 39.33 to 51.00 & 27.33 & 42.33 \\
Qwen2.5 7B & Bias aware & 56.22 & 52.89 to 59.56 & 57.67 & 52.00 to 63.00 & 40.33 & 31.67 \\
Qwen2.5 7B & Few shot & 55.78 & 52.56 to 59.22 & 55.67 & 50.00 to 61.33 & 39.00 & 33.67 \\
Qwen2.5 7B & Zero shot & 56.33 & 53.11 to 59.44 & 57.33 & 51.67 to 62.67 & 40.33 & 32.00 \\
\bottomrule
\end{tabular}%
}
\par\vspace{1.5mm}
\footnotesize{Values are percentages. CC denotes counterfactual consistency. Correct CC requires all persona variants from the same case to be answered correctly. NSE denotes narrative sensitivity error.}
\end{table*}

The main pattern changes once the 7B model is included. Scaling from 1.5B to 3B mainly improves accuracy, while the robustness gains are smaller and uneven. Scaling to 7B improves both accuracy and case level stability. Its best accuracy is 56.33 percent, and its counterfactual consistency reaches 57.67 percent under bias aware prompting. In practical terms, larger instruction tuned models reduce narrative instability, although they do not remove it.

The absolute accuracy values remain moderate even for 7B. For this reason, the models should not be interpreted as clinically ready systems. The accuracy likely reflects the difficulty of the medical question answering task, the compact to mid sized nature of the models, and the absence of medical domain specialization. The value of the experiment is the controlled comparison between accuracy and robustness, not validation for clinical use.

\begin{figure*}[t]
\centering
\includegraphics[width=0.92\textwidth]{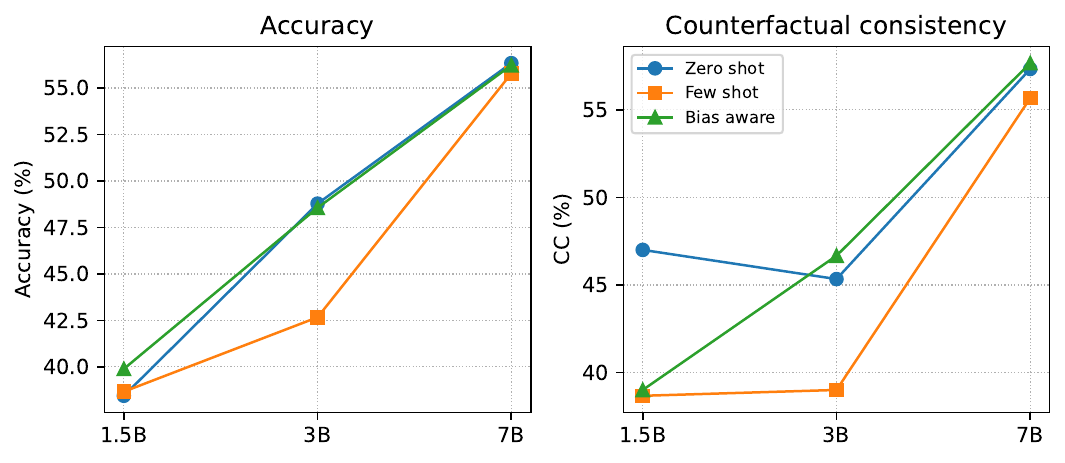}
\caption{Accuracy and counterfactual consistency across Qwen2.5 model scales under zero shot, few shot, and bias aware prompting. Performance generally improves with model scale, with the largest gains appearing at 7B.}
\label{fig:acc_cc}
\end{figure*}

Fig.~\ref{fig:acc_cc} makes the scale effect visible. Accuracy goes up steadily from 1.5B to 7B for all prompts. Counterfactual consistency also gets better at 7B, but the way prompts change is not exactly the same. Because of this, average correctness and case-level stability should still be reported separately.

\subsection{Narrative Sensitivity}
\begin{figure}[t]
\centering
\includegraphics[width=\columnwidth]{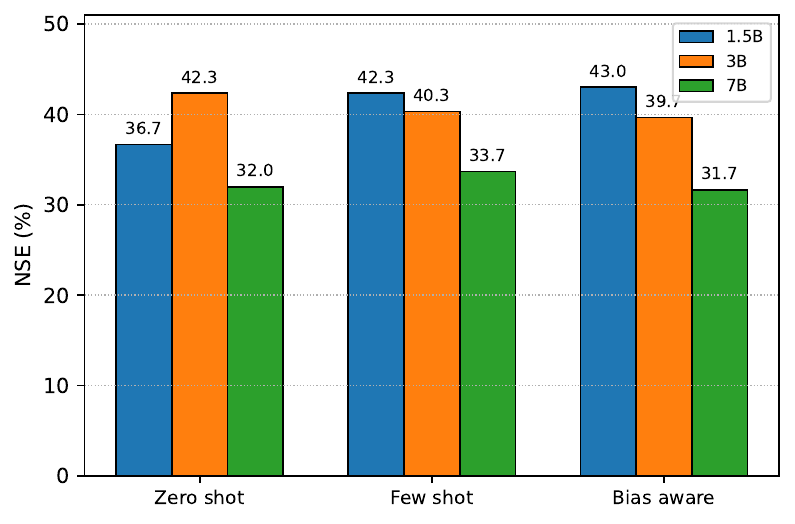}
\caption{Narrative sensitivity error across models and prompt settings. Lower values indicate fewer cases where matched narratives lead to mixed correct and incorrect answers.}
\label{fig:nse}
\end{figure}

Fig.~\ref{fig:nse} reports narrative sensitivity error. Lower values indicate fewer cases where one persona variant is correct and another equivalent variant is wrong. The 7B model has the lowest NSE in every prompt setting, reaching 31.67 percent with the bias aware prompt. Even so, nearly one third of case groups remain narrative sensitive under the best setting.

This result is important because narrative sensitivity is a case level failure mode. It appears when the model has enough information to answer at least one version correctly, but does not preserve that answer across the other versions. In clinical decision support, this pattern is troubling because it suggests that the response may depend on how the patient expresses the case.

\subsection{Case Level Error Decomposition}
Table IV summarizes the case-level error decomposition. To understand the consistency results, we break each case into four outcome types. Stable correct means every persona variant gets the right answer. Stable wrong means all variants get the same wrong answer. Narrative sensitive means at least one variant is correct and at least one is wrong. Unstable wrong means the model gives different answers across variants, but none are correct. This breakdown helps us tell apart missing medical knowledge from instability caused by the narrative.

\begin{table}[t]
\centering
\footnotesize
\caption{Case Level Outcome Decomposition}
\label{tab:decomp}
\resizebox{\columnwidth}{!}{%
\begin{tabular}{llrrrr}
\toprule
Model & Prompt & Stable Correct & Stable Wrong & Narrative Sensitive & Unstable Wrong \\
\midrule
Qwen2.5 1.5B & Bias aware & 17.33 & 21.67 & 43.00 & 18.00 \\
Qwen2.5 1.5B & Few shot & 18.00 & 20.67 & 42.33 & 19.00 \\
Qwen2.5 1.5B & Zero shot & 19.67 & 27.33 & 36.67 & 16.33 \\
Qwen2.5 3B & Bias aware & 28.00 & 18.67 & 39.67 & 13.67 \\
Qwen2.5 3B & Few shot & 22.67 & 16.33 & 40.33 & 20.67 \\
Qwen2.5 3B & Zero shot & 27.33 & 18.00 & 42.33 & 12.33 \\
Qwen2.5 7B & Bias aware & 40.33 & 17.33 & 31.67 & 10.67 \\
Qwen2.5 7B & Few shot & 39.00 & 16.67 & 33.67 & 10.67 \\
Qwen2.5 7B & Zero shot & 40.33 & 17.00 & 32.00 & 10.67 \\
\bottomrule
\end{tabular}%
}
\par\vspace{1.5mm}
\footnotesize{Values are percentages of 300 cases for each model and prompt setting.}
\end{table}

The breakdown explains why consistency in correct answers is still low. The 7B model raises the stable correct rate to about 40 percent, while the 3B model stays below 30 percent. However, cases that depend on the narrative are still common. So, even though the stronger model finds the right answer more often, it still struggles to keep that answer across all similar cases.

\subsection{Persona Level Accuracy}
Persona-level accuracy provides a descriptive view of where answer shifts occur. Across models, accuracy varied across alpha, beta, and gamma narratives, which supports the need for grouped case analysis rather than relying only on aggregate accuracy. Since persona-level differences were not the primary evaluation target, we report them as descriptive evidence only and focus the main analysis on case-level consistency, correct consistency, and narrative sensitivity error.

\subsection{Statistical Analysis}
McNemar exact tests confirm that larger Qwen2.5 models improve accuracy. Table~\ref{tab:mcnemar} reports paired tests for 1.5B versus 3B and 3B versus 7B. The 7B model is significantly more accurate than 3B in all prompt settings.

\begin{table}[t]
\centering
\footnotesize
\caption{Paired McNemar Tests for Model Scale}
\label{tab:mcnemar}
\resizebox{\columnwidth}{!}{%
\begin{tabular}{llrrr}
\toprule
Comparison & Prompt & A correct B wrong & A wrong B correct & $p$ value \\
\midrule
1.5B vs 3B & Bias aware & 124 & 202 & 0.000018 \\
1.5B vs 3B & Few shot & 134 & 170 & 0.044531 \\
1.5B vs 3B & Zero shot & 113 & 206 & $<0.000001$ \\
3B vs 7B & Bias aware & 107 & 176 & 0.000049 \\
3B vs 7B & Few shot & 90 & 208 & $<0.000001$ \\
3B vs 7B & Zero shot & 106 & 174 & 0.000058 \\
\bottomrule
\end{tabular}%
}
\par\vspace{1.5mm}
\footnotesize{A and B denote the first and second model in the comparison. Counts are discordant paired outcomes on the same question, persona, and prompt setting.}
\end{table}

By comparison, bias aware prompting does not significantly improve accuracy over zero shot prompting. Table~\ref{tab:prompt_mcnemar} reports the paired prompt comparison. The results show that telling the model to ignore irrelevant narrative cues does not reliably improve its accuracy. 

\begin{table}[t]
\centering
\caption{Prompt Comparison: Zero Shot Versus Bias Aware}
\label{tab:prompt_mcnemar}
\resizebox{\columnwidth}{!}{%
\begin{tabular}{llr}
\toprule
Model & Paired comparison & $p$ value \\
\midrule
Qwen2.5 1.5B & Zero shot vs. bias aware & 0.245891 \\
Qwen2.5 3B & Zero shot vs. bias aware & 0.880396 \\
Qwen2.5 7B & Zero shot vs. bias aware & 1.000000 \\
\bottomrule
\end{tabular}%
}
\par\vspace{1.5mm}
\footnotesize{McNemar exact tests compare paired correctness on the same case and persona rows.}
\end{table}

The bootstrap intervals confirm this finding. The 7B model has higher accuracy intervals than the smaller models in all prompt settings. Its counterfactual consistency intervals also increase, but narrative sensitivity remains. This suggests that scaling improves performance, but it does not fully solve the robustness problem.

\section{Discussion}
The main finding is straightforward. Scaling from 1.5B to 7B improves clinical answer accuracy and case level robustness. Qwen2.5 7B also improves correct consistency, which rises to 40.33 percent under zero shot and bias aware prompting. Even so, most cases are not stable correct across all three persona variants.

The prompt results are also mixed. Bias aware prompting slightly improves accuracy for Qwen2.5 1.5B and performs close to zero shot for Qwen2.5 3B and 7B. It does not significantly outperform zero shot in paired tests. More importantly, it does not remove narrative sensitivity error. This suggests that a short instruction to ignore irrelevant narrative cues is not enough for dependable clinical behavior.

From a clinical decision support perspective, the important issue is the split between correctness and stability. A model that is more accurate on average may still respond differently when the same case is written in another patient voice. In a real care setting, this could affect triage advice, diagnostic suggestions, or follow up guidance. For that reason, narrative robustness should be treated as a core evaluation target rather than a secondary check.

The equity concern is also central. SDoH cues can be clinically relevant when they affect risk, access, or care feasibility. The aim is not to take social context out of medical reasoning. Instead, we want to see if the model reacts too strongly to how a case is written when the clinical facts and answers stay the same. This difference matters for fair medical LLM evaluation.

The results also have an impact on how benchmarks are designed. Medical LLM reports should show both average accuracy and case-grouped robustness metrics, since the same overall score can hide very different levels of stability across patient narratives.

\section{Limitations and Reproducibility}
This study has some limitations. The experiment uses multiple choice questions instead of open-ended clinical support. This approach makes the evaluation easier to repeat, but it does not capture the full process of diagnostic reasoning or treatment planning. The dataset also relies on controlled persona variants. While these help isolate the patient’s voice, real clinical conversations are often more complex and may include missing information, mixed languages, or several rounds of clarification. A key limitation is that clinical equivalence across persona variants was taken from the dataset structure and fixed answer key rather than independently verified by physicians. Future work will include clinician review of matched variants to confirm that no clinically meaningful information changes across narratives.

The model set is limited to three Qwen2.5 instruction tuned models. This supports a controlled within-family scale comparison and helps isolate whether narrative variation affects answers as model size changes. However, the findings should not be generalized to all general-purpose or medical-domain LLMs. Future work will apply the same evaluation protocol to larger open models, proprietary systems where allowed, and clinically tuned models. The bias aware prompt is also simple by design. Stronger mitigation may require model training, calibration, retrieval safeguards, or clinician guided review. The absolute accuracy values are low, so the models should be treated as research systems rather than clinically usable systems.

The evaluation was set up to be reproducible. The dataset is public, the sampled identifiers use seed 42, and the models are available on Hugging Face. The code saves the parsed answer, gold answer, persona label, prompt condition, and case identifier for each row. This setup makes it possible to trace narrative-sensitive cases back to their matched variants, rather than relying only on overall scores.

\section{Conclusion}
This paper evaluated SDoH aware narrative anchoring bias in compact to mid sized LLMs using a counterfactual medical question answering setup. Out of 300 clinical cases and 8,100 model responses, Qwen2.5 7B was more accurate and robust than Qwen2.5 3B. However, even the stronger model was still affected by changes in how patient stories were told. Using bias-aware prompts did not make the models more accurate compared to zero-shot prompts. These results suggest that reliable clinical decision support needs more than just measuring average accuracy. Reports should also show if answers stay consistent when similar cases are described in different ways. Future research should apply this method to stronger open models, medical-specific models, patient stories in different languages, open-ended clinical reasoning, and safety reviews by clinicians.

\balance


\begin{thebibliography}{00}
\scriptsize
\setlength{\itemsep}{-1pt}
\setlength{\parsep}{0pt}

\bibitem{singhal2023clinical} K. Singhal \textit{et al.}, “Large language models encode clinical knowledge,” \textit{Nature}, vol. 620, July 2023, doi: 10.1038/s41586-023-06291-2.

\bibitem{kanjee2024limitations} P. Hager \textit{et al.}, “Evaluation and mitigation of the limitations of large language models in clinical decision-making,” \textit{Nature Medicine}, vol. 30, pp. 1–10, July 2024, doi: 10.1038/s41591-024-03097-1.

\bibitem{jin2021medqa} D. Jin, E. Pan, N. Oufattole, W.-H. Weng, H. Fang, and P. Szolovits, “What Disease Does This Patient Have? A Large-Scale Open Domain Question Answering Dataset from Medical Exams,” \textit{Applied Sciences}, vol. 11, no. 14, p. 6421, July 2021, doi: 10.3390/app11146421.

\bibitem{pal2022medmcqa} A. Pal, L. K. Umapathi, and M. Sankarasubbu, “MedMCQA: A Large-scale Multi-Subject Multi-Choice Dataset for Medical domain Question Answering,” \textit{PMLR}, pp. 248–260, Apr. 2022, Accessed: July 29, 2026. [Online]. Available: https://proceedings.mlr.press/v174/pal22a.html

\bibitem{jin2019pubmedqa} Q. Jin, B. Dhingra, Z. Liu, W. W. Cohen, and X. Lu, “PubMedQA: A Dataset for Biomedical Research Question Answering,” \textit{Empirical Methods in Natural Language Processing}, Sept. 2019, doi: 10.18653/v1/d19-1259.

\bibitem{lee2020biobert} J. Lee \textit{et al.}, “BioBERT: a pre-trained biomedical language representation model for biomedical text mining,” \textit{Bioinformatics}, vol. 36, no. 4, Sept. 2019, doi: 10.1093/bioinformatics/btz682.

\bibitem{yang2022gatortron} X. Yang \textit{et al.}, “A large language model for electronic health records,” \textit{npj Digital Medicine}, vol. 5, no. 1, pp. 1–9, Dec. 2022, doi: 10.1038/s41746-022-00742-2.

\bibitem{keloth2025sdoh} V. K. Keloth \textit{et al.}, “Social determinants of health extraction from clinical notes across institutions using large language models,” \textit{npj Digital Medicine}, vol. 8, no. 1, May 2025, doi: 10.1038/s41746-025-01645-8.

\bibitem{guevara2024sdoh} M. Guevara \textit{et al.}, “Large language models to identify social determinants of health in electronic health records,” \textit{npj Digital Medicine}, vol. 7, no. 1, pp. 1–14, Jan. 2024, doi: 10.1038/s41746-023-00970-0.

\bibitem{omiye2023race} J. A. Omiye, J. C. Lester, S. Spichak, V. Rotemberg, and R. Daneshjou, “Large language models propagate race-based medicine,” \textit{npj Digital Medicine}, vol. 6, no. 1, pp. 1–4, Oct. 2023, doi: 10.1038/s41746-023-00939-z.

\bibitem{zack2025sociodemographic} M. Omar \textit{et al.}, “Sociodemographic biases in medical decision making by large language models,” \textit{Nature Medicine}, Apr. 2025, doi: 10.1038/s41591-025-03626-6.

\bibitem{chen2023fairness} R. J. Chen \textit{et al.}, “Algorithmic fairness in artificial intelligence for medicine and healthcare,” \textit{Nature Biomedical Engineering}, vol. 7, no. 6, pp. 719–742, June 2023, doi: 10.1038/s41551-023-01056-8.

\bibitem{obermeyer2019racial} Z. Obermeyer, B. Powers, C. Vogeli, and S. Mullainathan, “Dissecting Racial Bias in an Algorithm Used to Manage the Health of Populations,” \textit{Science}, vol. 366, no. 6464, pp. 447–453, Oct. 2019, doi: 10.1126/science.aax2342.

\bibitem{rajkomar2018fairness} A. Rajkomar, M. Hardt, M. D. Howell, G. Corrado, and M. H. Chin, “Ensuring Fairness in Machine Learning to Advance Health Equity,” \textit{Annals of Internal Medicine}, vol. 169, no. 12, p. 866, Dec. 2018, doi: 10.7326/m18-1990.

\bibitem{bedi2025jama} S. Bedi \textit{et al.}, “Testing and Evaluation of Health Care Applications of Large Language Models,” \textit{JAMA}, Oct. 2024, doi: 10.1001/jama.2024.21700.

\bibitem{wang2025csedb} S. Wang \textit{et al.}, “A novel evaluation benchmark for medical LLMs illuminating safety and effectiveness in clinical domains,” \textit{npj Digital Medicine}, vol. 9, no. 1, Dec. 2025, doi: 10.1038/s41746-025-02277-8.

\bibitem{chugh2026dataset} P. Singh, “narrativeshield-sdoh-medqa,” huggingface, accessed 2026. [Online]. \url{https://huggingface.co/datasets/Prabhjotschugh/narrativeshield-sdoh-medqa}

\bibitem{qwen2025model} Qwen Team, ``Qwen2.5 model collection,'' Hugging Face Model Cards, accessed 2026. [Online]. Available: \url{https://huggingface.co/Qwen}

\end{thebibliography}
\end{document}